\documentclass{article}

\usepackage{microtype}
\usepackage{graphicx}
\usepackage{subfigure}
\usepackage{booktabs} 

\usepackage{amsmath}
\usepackage{amsthm}
\usepackage{url}
\usepackage{graphicx}
\usepackage[colorinlistoftodos]{todonotes}
\usepackage[colorlinks=true, allcolors=blue]{hyperref}
\usepackage{algorithm, algorithmic}

\usepackage{caption}
\usepackage{csvsimple}
\usepackage{longtable}
\usepackage{tabu}
\usepackage{authblk}

\makeatletter
\def\BState{\State\hskip-\ALG@thistlm}
\makeatother

\theoremstyle{definition}
\newtheorem{definition}{Definition}[section]

\usepackage{hyperref}

\usepackage[accepted]{icml2018}

\icmltitlerunning{MAGIX: Model Agnostic Globally Interpretable Explanations}

\begin{document}

\twocolumn[
\icmltitle{MAGIX: Model Agnostic Globally Interpretable Explanations}



\icmlsetsymbol{equal}{*}

\begin{icmlauthorlist}
\icmlauthor{Nikaash Puri (nikpuri@adobe.com)}{adobe} 
\icmlauthor{Piyush Gupta (piygupta@adobe.com)}{adobe} 
\icmlauthor{Pratiksha Agarwal (pratagar@adobe.com)}{adobe} 
\icmlauthor{Sukriti Verma (sukverma@adobe.com)}{adobe} 
\icmlauthor{Balaji Krishnamurthy (kbalaji@adobe.com)}{adobe}

\end{icmlauthorlist}

\icmlaffiliation{adobe}{Adobe Systems}

\icmlkeywords{Model Interpretation, Black box model understanding, Machine Learning, ICML}

\vskip 0.3in
]



\printAffiliationsAndNotice{\icmlEqualContribution} 

\begin{abstract}
Explaining the behavior of a black box machine learning model at the instance level is useful for building trust. However, it is also important to understand how the model behaves globally. Such an understanding provides insight into both the data on which the model was trained and the patterns that it learned. We present here an approach that learns if-then rules to globally explain the behavior of black box machine learning models that have been used to solve classification problems. The approach works by first extracting conditions that were important at the instance level and then evolving rules through a genetic algorithm with an appropriate fitness function. Collectively, these rules represent the patterns followed by the model for decisioning and are useful for understanding its behavior. We demonstrate the validity and usefulness of the approach by interpreting black box models created using publicly available data sets as well as a private digital marketing data set.
%
%
%
%
\end{abstract}

\section{Introduction}

Machine Learning and Artificial Intelligence have a large number of applications today. These include but are not restricted to applications in medicine, finance, imaging, operations in e-commerce, audio and so on. In such a situation, it becomes more important than ever to understand and interpret these algorithms. 
%
%

Broadly, when classified on the basis of complexity, there are two types of machine learning algorithms. The first are based on relatively straightforward formulations and are interpretable to a large extent. Under this category we have algorithms such as Linear Regression, Logistic Regression, Decision Trees and similar approaches. The main advantage of such algorithms is that they are simple and hence easy to interpret. For instance, by looking at the weights learned by a Linear Regression model, it is possible to determine the relative importance of different features used by the model. Hence, decisions made by such models are relatively easier to understand. 
%
%
%
%

The second type of algorithms are based on more complex formulations that are able to represent non-linear functions and higher order feature interactions. Popular among these are Neural Networks, Random Forests, Gradient Boosted Trees and various types of ensembles. One of the key characteristics of these approaches is that they are able to model very complex patterns and hence achieve higher accuracy on most data sets than their simpler counterparts. However, the cost of this gain in accuracy is loss of model interpretability. Neural Networks, for instance often have several hidden layers with different activations and dropout. Random Forests can have thousands of trees, and the final decision is a function of the combination of the individual predictions made by these trees. 

Model Interpretation is important in several industry verticals. It's importance has been discussed in detail in [\cite{lipton2016mythos}, \cite{ribeiro2016model}, \cite{doshi2017towards}] in the fields of medicine, finance, etc. However, an application that has not been covered in sufficient depth previously is that in Digital Marketing. Marketers spend a significant amount of money on personalization algorithms and platform to optimize content delivery. In such a scenario, it is important to have model interpretation techniques that can explain the behavior of the complex personalization algorithm to the marketer. Such a technique would increase the transparency and trust of the personalization platform. The approach we describe will be available to marketers as a part of a leading digital marketing suite of products.

We present here an approach that is global, model agnostic and that explains model behavior in an easy to understand way. Our approach learns rules that explain the behavior of a classification model. Each rule is independent, and is of the form \textit{If $C_1$ AND $C_2$ AND.... Then Predict Class $K$}. Here, $C_i$ refers to a specific condition such as '$15<age<25$' and $K$ is some class in the data set. Our approach is similar in objective to [\cite{lakkaraju2017interpretable}, \cite{bastani2017interpreting}, \cite{kim2017human}]. We demonstrate the results of our approach on data sets from medicine, finance and Digital Marketing. 
%
%
%
%
%
\section{Literature Survey}
There has been a variety of work in the field of model interpretation. The types of approaches applied can be characterized along several dimensions. They are:
\begin{enumerate}
\item The complexity of the response function that we want to explain. These include simple linear functions such as the ones learned by linear regression algorithms. Such functions are the most straightforward to interpret. The advantage of such functions is that it is easy to see what effect (both magnitude and direction) a particular attribute would have on the output variable. Non linear functions on the other hand involve complex combinations of the input features and are hence much harder to interpret. The advantage of representing model interpretations in the form of rule sets is that they can explain non-linear functions as well.  
%
%
\item Scope of interpretation. Some approaches allow global interpretation of machine learning models([\cite{lakkaraju2017interpretable}, \cite{bastani2017interpreting}, \cite{kim2017human}, \cite{shrikumar2016not}]), while others focus on instance level or local explanations([\cite{ribeiro2016should}, \cite{fong2017interpretable}, \cite{lundberg2017unified}, \cite{robnik2008explaining}, \cite{kononenko2010efficient}, ]). By local explanations, we mean explaining model behavior in a limited region of the input space. 
\item Whether the approach depends on the model. There are two types of approaches under this category. The first are model specific, i.e. those approaches that are designed to exploit model specific properties in constructing explanations[\cite{foerster2017input}, \cite{selvaraju2016grad}, \cite{goyal2016towards}, \cite{sundararajan2017axiomatic}, \cite{setiono2004approach}, \cite{shrikumar2016not}]. The second are model agnostic, i.e. those approaches that do not leverage underlying details of the machine learning model in constructing explanations[\cite{lakkaraju2017interpretable}, \cite{bastani2017interpreting}, \cite{kim2017human}, \cite{koh2017understanding}, \cite{henelius2017interpreting}, \cite{phillips2017interpretable}, \cite{fong2017interpretable}]. Rather, they treat the model as a black box and rely on the predict function of the model.  
\end{enumerate}

Techniques such as partial dependence plots, residual analysis and generalized additive models have been used to understand model behavior as described in \cite{oreily_model_interpretation}. These techniques are useful for models that learn monotonic response functions. However, complex models such as neural networks and gradient boosted trees learn non-monotonic response functions. In such functions, changes in the input variables in the same direction can lead to the response variable changing in different directions. For example, in a loan rejection model, the age increasing from 20 to 25 may decrease the risk of loan rejection, however the age increasing from 60 to 65 might produce the opposite effect and increase the risk of loan rejection. Interpreting the behavior of such models at a global level is a non-trivial problem. 

\cite{tan2017detecting} and \cite{tan2018transparent} investigate how model distillation can be used to distill complex models into into models that are transparent or interpretable in some sense.

\cite{bastani2017interpreting} have used a surrogate model approach where they extract a decision tree that represents model behavior. The problem with this approach is that each path to a leaf node goes through the first attribute that the tree splits on. The result being that each rule includes a condition involving this attribute. This holds recursively such that the rules derived from the left subtree would have two common attributes and so on. Further, the tree itself can often be several levels deep, leading to complex paths that are not easy to understand. Hence, we have focused our work on output in the form of rule lists as described in \cite{letham2015interpretable} and \cite{lakkaraju2016interpretable}. The latter describe a Joint Framework for Description and Prediction in the form of interpretable decision sets. They show that decision sets are more comprehensible to humans than decision lists because rules apply independently. The output of our approach is in the form of a decision set, where each rule applies independently of the others.  
%
%
%
%
%

The global black box model interpretation approaches discussed in [\cite{lakkaraju2017interpretable}, \cite{lakkaraju2016interpretable}] use the Apriori algorithm \cite{agrawal1996fast} to generate conditions. The problem with this approach is that it generates conditions that have high frequency in the data set and not necessarily those that are useful for distinguishing between classes. In our approach we have used the LIME algorithm proposed by \cite{ribeiro2016should} to learn conditions that are important to explain the classification of certain instances. Rules are learned by a genetic algorithm that tries different combinations of conditions to optimize a fitness function. Hence, our approach can be thought of as an extension to LIME that explains model behavior globally in the form of independent if then rules. 

\section{Problem Description}
Our method takes as input a Data-set $\mathcal{D}$ and a model $\mathcal{M}$ trained using that Data-set.\\
The dataset $\mathcal{D} = \{(x_i, y_i)\}_{i=1}^N$, where each row consists of an instance $x_i = \{x_i^1, x_i^2,..,x_i^p\}$ having fixed $p$ columns and a class label $y_i \in \textbf{Y} = \{Y_1,..,Y_k\}$, the set of $k$ classes.\\
The classification model $\mathcal{M}$ has a function \textit{predict-proba}($x_i$) that assigns a $1\times k$-vector of probabilities to each $x_i \in \mathcal{D}$. Each element in the vector represents the probability that the instance belongs to the corresponding class.\\
The algorithm outputs the set of rules for each class $Y_j$. Each rule is a conjunction of conditions and a condition is a set of constraints on values of one feature column.

\section{Notations and Definitions}
Please refer to Table \ref{table:notations-and-definitions} for a reference of the notations and definitions used in the paper. 

\begin{table}
\caption{Notations and Definitions Used}
\label{table:notations-and-definitions}
\begin{tabular}{|p{.07\textwidth}|p{.25\textwidth}|p{.1\textwidth}|}
\hline
Symbol                 & Definition                                                                                                                                                                                                                                                        & Example                                                                \\ \hline
\textbf{X}             & training data, consisting of instances                                                                                                                                                                                                                            & {[}(1, John, 25), (2, Jane, 30) ...{]}                                  \\ \hline
\textbf{Y}             & set of predicted classes                                                                                                                                                                                                                                          & {[}approve, reject{]}                                                  \\ \hline
$x$                     & Particular instance in the training set                                                                                                                                                                                                                           & (1, John, 25)                                                          \\ \hline
$y$                     & Particular predicted class                                                                                                                                                                                                                                        & approve                                                                \\ \hline
$c_i$                     & Condition, which is a combination of an attribute and a range of values the attribute can take                                                                                                                                                                               & age\textless30, state$\in$\{New York, Florida\}                                                         \\ \hline
$R_i$                     & A classification rule which classifies an instance to belong to its associated class if a predicate consisting of a conjunction(AND) of conditions holds.                                                                                                                                                                                                                             & IF age\textless30 and salary\textgreater100 and state=New York Then Predict Class approve \\ \hline
$C_i$                     & Set of conditions for rule $R_i$                                                                                                                                                                                                                                  &\{age\textless30, salary\textgreater100, state=New York\}                            \\ \hline
$y_i$                     & Associated target class for rule $R_i$                                                                                                                                                                                                                                & Predict Class: approve                                                  \\ \hline
cover($R_i$)              & Set of instances rule $R_i$ covers in the training data. These are instances for which the conditions of $R_i$ are true.                                                                                                                                             & \text{\{$x_1, x_2$\}}                                                             \\ \hline
correct-cover($R_i$)      & Set of instances for which the rule's class prediction $y_i$ agrees with the model's prediction.
Formally, it is the set $\{ x | x\in$ cover($R_i$) and $y_i=\mathcal{M}(x_i)\}$ where $\mathcal{M}(x_i)$ is the class predicted by the classifier for the instance $x_i$ & \text{\{$x_1$\}}                                                                \\ \hline
incorrect-cover($R_i$) & Set of instances that are incorrectly covered by $R_i$                                                                                                                                                                                                            & \text{\{$x_2$\}}                                                                \\ \hline
\textbf{R}             & Rule set. Consists of a set of rules of the form $R_i$.                                                                                                                                                                                                           & \text{\{$R_1$, $R_2$, ...\}}                                                \\ \hline
correct-cover(\textbf{R})       & Correct cover of rule set \textbf{R}. It is defined as the union of the correct covers of $R_i$ for each rule in \textbf{R}                                                                                         & \text{\{$x_1, x_3$\}}                                                              \\ \hline
cover(\textbf{R})               & Cover of rule set \textbf{R}. It is defined as the union of the covers of $R_i$ for each rule in \textbf{R}                                                                                                         & \text{\{$x_1, x_2, x_3$\}}                                                          \\ \hline
\end{tabular}
\end{table}

\begin{definition}{Rule Precision:}\label{def:rule-precision}
The precision of $R_i$ is the ratio of number of instances in correct-cover($R_i$) to the number of instances in cover($R_i$).
\[Precision(R_i) = \frac{\textit{len}(correct\text{-}cover(R_i))}{\textit{len}(cover(R_i))} \]
\end{definition}

\begin{definition}{Rule Length:}
The length of $R_i$ is the cardinality of the precondition set for the rule $C_i$.
\[ Length(R_i) = \text{number of conditions in $R_i$} \]
\end{definition}

\begin{definition}{Rule Class Coverage:}\label{def:rule-class-coverage}
The class coverage of $R_i$ is the ratio of  number of instances in correct-cover($R_i$) to the number of instances in the training set that have been predicted by the classifier to have label $y_i$.
\end{definition}

\begin{definition}{Rule Mutual-Information (\textbf{RMI}):}\label{def:rule-mutual-information}
The \textit{Rule Mutual-Information} or \textit{RMI} of a rule $R_i$ captures its mutual dependence with the class predicted. It is derived using the Mutual Information($MI$) of the contingency table \ref{table:rule-mutual-information} as follows -
\[ RMI(R_i) = \left\{ \begin{array}{rl} MI & when\;n_{i1} \geq n_{i2} n_{i3}/n_{i4} \\
-1\times MI & otherwise \end{array}\right. \]
\end{definition}

\begin{table}
\caption{Contingency table used to compute \textit{RMI} of rule $R_i$}
\label{table:rule-mutual-information}
\begin{tabular}{|p{.08\textwidth}|p{.15\textwidth}|p{.17\textwidth}|}
\hline
Rule/Class & Predicted-class  & Other class(es) \\ \hline
$R_i$ & $n_{i1}$ = Cardinality of correct-cover($R_i$) & $n_{i2}$ = Cardinality of incorrect-cover($R_i$) \\ \hline
NOT $R_i$ & $n_{i3}$ = Count of instances with same label as the class predicted but not covered by $R_i$ & $n_{i4}$ = Count of instances not included in $R_i$ and with label different from the class predicted by $R_i$ \\ \hline
%
%
\end{tabular}
\end{table}

\begin{definition}{Rule Set Precision:}\label{def:rule-set-precision}
The precision of \textbf{R} is the ratio of  number of instances in correct-cover(\textbf{R}) to the  number of instances in cover(\textbf{R}).
\end{definition}

\begin{definition}{Rule Set Class Coverage:}\label{def:rule-set-coverage}
The class coverage of \textbf{R} is the ratio of  number of instances in correct-cover(\textbf{R}) to the number of instances in the training set that have been predicted by the classifier to have label $y_i$ where $y_i$ is the class that the rules in \textbf{R} predict.
\end{definition}

\section{Approach}

Our approach is outlined in Algorithm \ref{alg:explain-model} and the subsequent sections explain each step in more detail. 
%
%
%
%

\begin{algorithm}[h]
\caption{EXPLAIN-MODEL}\label{alg:explain-model}
\begin{algorithmic}[1]
\STATE $\textbf{X} \gets \text{training instances}$
\STATE $\textbf{M} \gets \text{classification model}$
\STATE $\textbf{Y} \gets \text{set of output classes}$
\STATE $\textbf{R} \gets \text{Map$<$string, set()$>$}$
\STATE $\textit{PreProcessInputData}(\textbf{X}, \textbf{Y})$
\FOR{$y_i$ in \textbf{Y}}
\STATE $\text{conditions} \gets \textit{GenInstConds}(\textbf{X}, \textbf{M}, y_i)$
\STATE $\text{classLevelRules} \gets \textit{LrnCls}(\textbf{X}, \textbf{M}, conditions, y_i)$
\STATE $\text{classLevelRules} \gets \textit{Proc}(classLevelRules, y_i)$
\STATE $\textbf{R}[y_i] \gets \textit{Sort}(optimalRuleSubset, y_i)$
\ENDFOR
\STATE return \textbf{R}
\end{algorithmic}
\end{algorithm}

\subsection{Preprocessing the Input Data}\label{section:preprocess-input-data}

In this step we pre-process the input data to make each feature categorical. Concretely, for categorical features we leave them unmodified. For numerical features, we perform entropy based binning to split the feature values into discrete bins [\cite{ribeiro2016should}]. For instance, if the attribute age takes values from 10 to 85 (inclusive), then our binning step could produce the ranges   
%
%
'$10\le age<25$', '$25\le age<60$' and '$60\le age\leq 85$'. Post this step, the input data comprises only of categorical features. Further, we split the original data set into training and testing data sets. 

These preprocessing steps correspond to the \textit{PreProcessInputData}(\textbf{X}, \textbf{Y}) function in Algorithm \ref{alg:explain-model}.

\begin{figure}
  \includegraphics[width=8.0cm]{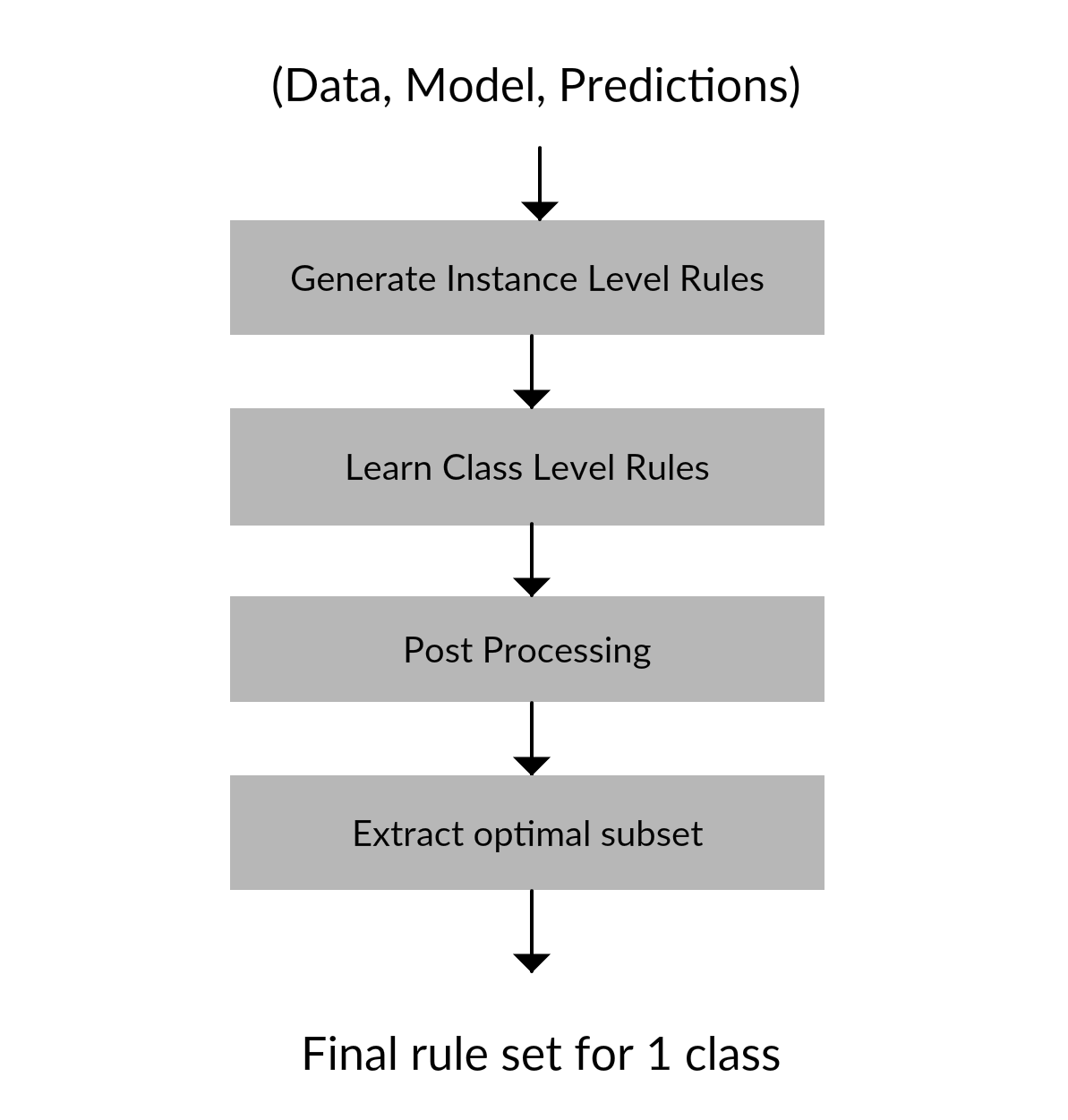}
  \caption{Framework to get rules for one class}
  \label{fig:algo_fig}
\end{figure}

\subsection{Generating Instance Level Conditions}

The algorithm iterates over each instance in the training data. For each instance there is a particular class that the instance has been classified into. We compute the marginal contribution of each feature-value used by the model in arriving at this classification. We have used the approach described by \cite{ribeiro2016should} to compute instance level marginal contributions. This corresponds to the procedure \textit{Marg($x_i$, M)} in Algorithm \ref{alg:generate-instance-level-conditions}. The approach perturbs the training instance in question, and trains a locally faithful linear model in the locality of the instance under consideration. The weights of the different features then approximate the marginal contribution values. The output of this step is a list of conditions. Each condition consists of a single feature and a value. The complete algorithm is outlined in Algorithm \ref{alg:generate-instance-level-conditions}. 
This corresponds to the \textit{GenInstConds} function in Algorithm \ref{alg:explain-model}.
%
%
\begin{algorithm}[h]
\caption{GEN-INST-CONDS}\label{alg:generate-instance-level-conditions}
\begin{algorithmic}[1]
\STATE $\textbf{X} \gets \text{training instances}$
\STATE $\textbf{M} \gets \text{classification model}$
\STATE $\text{not-yet-covered} \gets \textbf{X}$
\STATE $\text{conditions} \gets \text{set()}$
\FOR{$x_i$ in \textbf{X}}
\STATE $\text{instance-level-conditions} = \textit{Marg($x_i$, M)}$

\FOR{$cond$ in instance-level-conditions}
\STATE $\text{not-yet-covered} \gets \text{not-yet-covered} - \textit{CoveredInstances(cond)}$
\ENDFOR

\STATE $\text{conditions} \gets \text{conditions} \cup \text{instance-level-conditions}$

\IF {$\textit{length}\text{(not-yet-covered)} = \text{0}$}
\STATE \textbf{break};

\ENDIF
\ENDFOR
\STATE \text{return conditions};

\end{algorithmic}
\end{algorithm}

\subsection{Learning Rules From Conditions}\label{subsection:learning-rules-from-conditions}

The output of the previous step of generating instance level conditions is a list of conditions that are important at the instance level. For each class, we have a set of conditions that were important in classifying instances of that class. We want to learn rules for that class. Each rule has an associated coverage and precision as defined in definitions \ref{def:rule-class-coverage} and \ref{def:rule-precision} respectively. Hence, the problem is the following. How can we build rules having high precision and coverage, from these base conditions? This could be done in a combinatorially prohibitive manner. For instance, we could first evaluate all rules of length 1, then length 2 and so on. However, if the output of the previous step gives N conditions, then the complexity of this algorithm would be $2^N$. This is not computationally feasible for larger data sets. 

We need an approach that can learn optimal combinations of conditions. And we want the notion of optimality to be abstracted away from the particular approach. Concretely, we want an algorithm that can take as input a notion of optimality and subsequently combine conditions to generate rules that are optimal. Another point that needs to be considered is that if there are any categorical variables in the data then we want the rules to allow these categorical variables to take more than one value. For example, if we had a categorical variable called country and if '$10\leq age < 25$', '$country = US$', and '$country = India$' are conditions for class 2, then one candidate rule could be 'If $10\leq age < 25$ AND $country \in \{US, India\}$ THEN Predict Class 2'. Therefore, the combinations that we need are not only 'AND's of conditions but also 'OR's within a condition involving a categorical variable so that it allows such a variable to take on multiple possible values.

Similar to the work done in \cite{fidelis2000discovering} and \cite{ghosh2010mining}, we use a Genetic Algorithm to learn such rules under the given conditions. The algorithm is run independently for each class. Hence, we are trying to learn class level rules from class level conditions. Each individual of the Genetic Algorithm represents a possible rule. For example, if, the number of conditions generated in the previous step (learning instance level conditions) was 100, then, each individual of our population would be a bit string of length 100. One example string might be '100100000....000'. The example individual described represents a rule of the form 'If condition1 AND condition4 THEN predict Class 1'. Each condition that involves a categorical variable has been allowed to take more than one value. The fitness function should capture several requirements:
\begin{enumerate}
\item \textbf{Precision and Coverage:} Rule Mutual-Information (RMI) as defined in definition \ref{def:rule-mutual-information}, captures this notion of simultaneously trying to optimize for both precision and coverage. The rules having a high RMI are neither highly precise with low coverage nor overly generic and imprecise with high coverage.
\item \textbf{Length:} We want the fitness function to add a length factor. The intuition here is that shorter rules are easier to interpret. And while it is true that rule length often inversely correlates with coverage, however we find that making this parameter explicit leads to shorter, more interpretable rules.
\item \textbf{Overlap:} It may happen that an instance is covered by one rule that says it should belong to Class 1 and another rule that says it should belong to Class 2. Clearly, only one of these is correct and we want to minimize the amount of ambiguity in our final rule set. Since we plan to go for rules that have high precision, meaning that they were correctly copying model behavior, the amount of overlap in the final rule set will be minimized as we optimize precision.
\end{enumerate}

Keeping all these in mind, we designed the following fitness function for our genetic algorithm:
\[F(ind) = RMI(R_i) - \frac{\textit{No. of active bits in ind}}{N}
\]

Here, \textit{ind} stands for individual represented by a bit string, $R_i$ is the rule that this individual represents and N is the length of the bit string.
A weight can be given to either of these terms as per the requirement of the user. We take our population to be 1200 individuals with a cross-over probability of 50\% and mutation probability set in such a manner that on average 2 bits of an individual are flipped whilst undergoing mutation. This gives a reasonable trade-off between exploration and exploitation. We initialized the population with individuals that have a high probability of being 'fit'. These are individuals with only one bit set in the bit string (such as, 1000, 0100, 0010 and 0001), followed by those with only two bits set (1100, 0110, 0101...) and so on and so forth until the entire population size i.e. 1200 individuals, is reached. We run the algorithm for 600 generations. All the individuals of the last generation are finally selected as our rule set for one class. \newline
Hence, the output of this step is an exhaustive rule set for each class. There will be several rules that do not add value to the set and need to be filtered out.

This step corresponds to the \textit{LrnCls} function in Algorithm \ref{alg:explain-model}.

\subsection{Post-processing Rules}

The Rules generated in the previous step include several redundant rules. To eliminate them, we sort the generated rules in descending order of precision. Then, for each rule $R_i$, we check whether correct-cover($R_i$) $\subseteq$ correct-cover($R$) and Precision($R_i$)$\le$Precision($R$), where $R$ is a more precise rule not yet removed. Then we do not consider $R_i$ for the next step. Otherwise, we retain this rule for consideration. 

We want to avoid getting spurious rules that do not represent the patterns that the black box model learnt. Such rules are possible due to Texas sharpshooter fallacy[\cite{texas_sharpshooter_fallacy}] and do not describe model's prediction on unseen data. The original data set is split into training and testing subsets. The rules are learned by observing model behavior on the training data set. Then, for each rule, the precision on the testing data set is calculated. If the precision is less than the baseline precision on the test data, the rule is discarded at this step. This helps by removing noisy rules that have a positive RMI in the training data set but low precision in the test data set. Baseline precision is the fraction of the instances in the test set for which the model predicts $y_i$, the class associated with the rule $R_i$.

The corresponding procedure in Algorithm \ref{alg:explain-model} is \textit{Proc}(\textbf{R}). 

\subsection{Sorting rules by Mutual Information}\label{section:sorting}
The previous section gives us a combination of 'OR's (a subset of rules) among the 'AND's (combining clauses into rules) generated by the Genetic Algorithm. Once we have the optimal set of rules for a class, we want to sort them so as to provide the user with the most relevant rules at the top and less relevant ones further down the list. This is done as follows. First, the rules are sorted in descending order of RMI. Then, the rule with the highest RMI is selected, and added it to the top of the list. Then, the rule with the next highest value of RMI is selected. The rule is compared to the list of already added rules. If it is similar to any of the rules already added, we discard this rule. Otherwise, it is added to the list of rules. The similarity is computed using the Jaccard Similarity measure between the list of instances covered by the two rules. If the Jaccard Similarity is greater than 50\%, the rules are deemed to be similar. This process is repeated till we have a desired number of rules. 

The corresponding procedure in Algorithm \ref{alg:explain-model} is \textit{Sort}(\textbf{R}).

\section{Results}\label{section:results}
We demonstrate our approach on four publicly available data sets and explain the results across various measures and dimensions to draw conclusions. These are the 'Iris' (\cite{fisher1936use}) data set, the 'Wisconsin Breast Cancer data set'(\cite{Lichman:2013}), the 'Banknote authentication data set'(\cite{Lichman:2013}) and the 'Car Evaluation Dataset'(\cite{Lichman:2013}). We train a random forest classifier (with 500 trees, min-samples-split=2, min-samples-leaf=1) on each dataset and interpret the models as a decision set of independent rules using algorithm \ref{alg:explain-model}. Table \ref{table:all-results} shows a sample of the rules obtained. 

\begin{table}
\caption{Sample of Results on Datasets}
\label{table:all-results}
\begin{tabular}{|p{.07\textwidth}|p{.26\textwidth}|p{.03\textwidth}|p{.03\textwidth}|}
\hline
Dataset       & Rule                                                                                                        & Prec. (\%) & Cov. (\%) \\ \hline
Iris          & petal-width \textless= 0.80 THEN Predict class: Iris-setosa                                                 & 100            & 100           \\ \hline
Iris          & petal-width \textgreater 1.85 THEN Predict class: Iris-virginica                                            & 100            & 62            \\ \hline
Iris          & 2.60 \textless petal-length \textless= 4.45 THEN Predict class: Iris-versicolor                                            & 100            & 62            \\ \hline

Breast Cancer & 0.50 \textless bare\_nuclei \textless= 1.50 AND mitoses \textless= 1.50 THEN Predict class: 2               & 98             & 83            \\ \hline
Breast Cancer & bare\_nuclei \textgreater 8.50 THEN Predict class: 4                                                        & 97             & 56            \\ \hline
Breast Cancer & clump\_thickness \textgreater 8.50 THEN Predict class: 4                                                    & 100            & 33            \\ \hline
Banknote      & variance \textgreater 2.39 THEN Predict class: 0                                                            & 100            & 52            \\ \hline
Banknote      & skewness \textgreater 9.62 THEN Predict class: 0                                                            & 100            & 16            \\ \hline
Banknote      & -2.80 \textless variance \textless= -0.40 AND -5.87 \textless entropy \textless= 1.00 THEN Predict class: 1 & 83             & 43            \\ \hline
Banknote      & -6.98 \textless skewness \textless= -5.51 THEN Predict class: 1 & 84             & 11            \\ \hline

Cars          & safety = low THEN Predict class: unacc                                                                      & 100            & 49            \\ \hline
Cars          & persons = more,4 AND safety = high AND buying = med THEN Predict class: acc                                 & 70             & 17            \\ \hline
Cars          & persons = 2 THEN Predict class: unacc                                 & 100             & 46            \\ \hline
Cars          & safety = high AND lug-boot = small AND persons = 4 AND buying = med,low AND maint = low,med THEN Predict class: good                                 & 75             & 20            \\ \hline

Digital Marketing          & CUMULATIVE-ACTION \textless= 0.50 AND MOB-targeting.mobile.displayHeight \textless= 300.00 THEN Predict class: 1                                 & 100             & 87.4            \\ \hline
Digital Marketing          & ENV-OperatingSystemVersion = Linux AND MOB-targeting.mobile.osAndroid = 1 THEN Predict class: 2                                 & 51.1             & 46            \\ \hline
Digital Marketing          & ENV-OperatingSystemVersion = Linux AND 597.50 \textless ENV-BrowserWidth \textless= 607.50 AND 567.00 \textless ENV-ScreenWidth \textless= 620.50 THEN Predict class: 2 & 84.6             & 22            \\ \hline
\end{tabular}
\end{table}

\subsection{Iris Dataset}

The data sets consists of 3 different types of irises’ (Setosa, Versicolour, and Virginica). It is a classification data set, having four features which are Sepal Length, Sepal Width, Petal Length and Petal Width. The three classes are Setosa, Versicolour, and Virginica. The Random Forest classifier accuracy was close to 98\%.

Table \ref{table:all-results} shows a sample of the rules extracted from a Random Forest model trained on this dataset using Algorithm \ref{alg:explain-model}. As an example consider the rule 'IF petal-width (cm) $<$= 0.80 THEN Predict class: Iris-setosa'. It has a precision of 100\%. This means that whenever the classifier saw a flower having petal width less than 0.80cm, it classified it into the class Setosa. A domain expert could now determine whether such a rule is reasonable in the real world, or is something that inadvertently crept into the data set and hence not likely to hold for real world applications. Further, a data scientist could analyze such rules to evaluate the extent to which they hold in the data set. 

\subsection{Wisconsin Breast Cancer Dataset}\label{section:results-wisconsin-breast-cancer}
This dataset was obtained from the University of Wisconsin Hospitals, Madison from Dr. William H. Wolberg \cite{Lichman:2013}. The Random Forest classifier accuracy was close to 98\%. 

The rules obtained for this data set offer some insight into both the workings of the classifier and the original data set. We discuss a couple of rules here. As we can see from Table \ref{table:all-results}, the rule 'IF clump-thickness greater than 8.50 THEN Predict class=4(malignant)' is a high precision rule. At this point a domain expert could examine this and similar rules to gain an understanding of patterns in the training data set. Further, if such rules are due to spurious introductions into the training data and not due to genuine patterns, then the model can be retrained with a more diverse data set to increase its generalization power. By looking at only instance level justifications, it is not easy to understand the general patterns that the model derived from the data. Hence, the aggregated view of model behavior that our approach produces is useful to understand how the model works.

\subsection{Banknote Authentication Dataset}
This dataset was extracted from images taken from genuine and forged banknote-like specimens \cite{Lichman:2013}. Wavelet Transform tool was used to extract features from these images. The dataset has 1372 instances, described by 4 attributes. The Random Forest classifier accuracy was close to 99\%. Table \ref{table:all-results} shows a sample of the rules obtained from the classifier. 

\subsection{Car Evaluation Dataset}
Each instance of this dataset describes a car \cite{Lichman:2013}. This dataset has 1728 instances, described by 6 attributes: buying cost, maintenance cost, number of doors, number of persons, size of lug boot, safety indication. All of the attributes are categorical taking values like high, medium, low and etc. The target variable indicates whether a particular instance i.e. a car is unacceptable, acceptable, good or very good. We trained a Random Forest classifier with 400 trees on the data set. The classifier accuracy was close to 96\%. The rules obtained are shown in Table \ref{table:all-results}. For instance, the rule 'safety=low THEN Predict class: unacceptable' has 100\% precision and makes sense intuitively. 

\subsection{Digital Marketing Dataset}
We have also tested our approach on several real life marketing datasets from a personalization platform of a large digital marketing company. The platform learns from user behavior and shows the right content to the right user with the objective of maximizing the user's likelihood of conversion. Conversion could correspond to clicking an ad, buying a product, etc.. The platform is driven by machine learning models such as Random Forests. With our approach, the decisioning logic of the black box models are made transparent and this leads to increased trust and provides insights to the marketers.

The data set from the platform consists of a set of users with their profiles. For each user, we have the piece of content that the platform decided to show that user. Further, we have a serialized form of the model used by the platform to perform the decisioning. Table \ref{table:all-results} shows a sample of the rules obtained. A set of such rules would help the marketer better understand the behavior of the digital marketing platform. 

Further, we have created another data set to help the marketer understand user behavior. The data set consists of a set of users with their profiles and whether they converted. It is important to state that we record whether the user converted regardless of the piece of content for which they converted. A model is then trained on this data set. Finally, we use MAGIX to interpret the model. The rules so obtained describe user segments that have either a high or a low propensity to convert. This could be useful for the marketer to understand how his end users behaved. 

We have validated the usefulness of the approach with several customers. Marketers will be able to login to the platform and view both types of reports (model explanation report and user behavior report).

\subsection{Validation}
After completion of the rule processing step, we have a set of rules for each class. The rules have been selected in a manner that optimizes for coverage and precision simultaneously. These metrics are useful for evaluating how well a single rule expresses model behavior. However, we need an approach for quantifying how well a collection of rules represents the model. In order to achieve this, we used the rules, discovered by our method, to build a predictive model. The logic for the model is as follows. If an instance is covered by more than one rule, then it is classified according to the most precise rule. Ties are settled randomly. Then, we use both the original model and our rule based model to make predictions on the testing data. Finally, we compare the predictions made by our rule based model to those made by the original model and compute the fraction of predictions that match. We call this the \textit{Imitation@K} metric. Here, K refers to the number of rules that the rule set contains. This is to ensure that a scheme does not obtain very high imitation accuracy by including a very large number of high precision and low coverage rules. Approaches that have a number of very high precision, low coverage rules would cover a fewer number of instances and hence receive a low score in the metric. Similarly, approaches that have high coverage, low precision rules would make a large number of mistakes and hence receive a low score in the metric. To achieve a good score, the set of rules must cover a large fraction of instances with a high precision.  

\begin{table}
\centering
\caption{Validation Results for different values of K}
\label{table:validation}
\begin{tabular}{|p{0.07\textwidth}|p{0.05\textwidth}|p{0.05\textwidth}|p{0.05\textwidth}|p{0.05\textwidth}|p{0.05\textwidth}|}
\hline
Dataset                 & Im@1 & Im@2 & Im@5 & Im@10 & Im@20 \\ \hline
Iris                    & 33.33	& 66.66 & 86.66 & 96.66 & 96.66 \\ \hline
Breast Cancer           & 56.42 & 58.57 & 63.57 & 93.57 & 95.71 \\ \hline
Banknote Authentication & 32.00 & 50.90 & 71.63 & 94.90 & 96.00 \\ \hline
Car Evaluation          & 24.56 & 53.75 & 86.12 & 90.17 & 89.01 \\ \hline
\end{tabular}
\end{table}

Table \ref{table:validation} shows the validation accuracies (in percentages) for the different data sets for different values of \textit{K}. As can be seen from the table, our approach generates rules that are able to imitate model behavior for a large fraction of the data set. This indicates that the set of rules is able to capture to a large extent the behavior of the model. It is also important to note that as we increase the value of \textit{K}, it is possible that imitation accuracy goes down. This is because the addition of rules could cause certain instances to be classified incorrectly by the rule based proxy model.  

\section{Conclusion}
We have presented here an approach that explain a black box model globally. Our experiments on different data sets demonstrate the usefulness of our approach. The potential applications of our approach are diverse. These include applications in medicine, finance and digital marketing platform solutions. Further, by understanding models trained on large data sets, we can extract patterns inherent in the original data. This gives us a useful way to understand large and complex data sets. We have also introduced the \textit{Imitation@K} metric that could be useful for comparing decision rule sets generated through different approaches. 

\section{Limitations of MAGIX}
Our approach only works with classification models. This is because the decision rule set may not be a good interpretable representation for a regression model, where we would like to present the output also as a function of features. Further, we have not applied the approach to image classification problems. While it is possible to use \cite{ribeiro2016model} to derive image regions that are important for classification at the instance level, it is not clear how to aggregate such regions to create global rules that are applicable for the entire training dataset. 

\section{Future Work}

The current version of the approach is based on classification problems. One possible direction of future research is to extend our approach to explaining models that predict continuous variables. Further, it would be useful to extend our approach to explain models that work with images. As model interpretation approaches become more prevalent, another important direction of research is to devise metrics that can compare two model interpretation approaches. We have proposed the \textit{Imitation@K} metric. However, this is only useful for comparing interpretation approaches that output a set of rules. A more general metric that can compare any two interpretation approaches would be useful.

\bibliography{main}
\bibliographystyle{icml2018}

\end{document}